\documentclass[11pt,a4paper]{article}
\usepackage[hyperref]{emnlp2020}
\usepackage{standalone}
\usepackage{mystuff-nlp}
\usepackage{times}
\usepackage{tikz}
\usepackage{url}
\usepackage{latexsym}
\usepackage{color}
\usepackage{float}
\usepackage{amsfonts,amssymb,amsmath,bm,bbm}
\usepackage{algorithmic}

\usepackage{graphicx,subfigure}
\usepackage{booktabs}
\usepackage{multirow}
\usepackage{verbatim}
\usepackage{enumitem,array}
\usepackage{setspace}
\usepackage{appendix}

\usepackage{arydshln}

\usepackage[titlenumbered,ruled]{algorithm2e}

\usepackage{microtype}

\newcommand{\example}[1]{`#1'}
\newcommand{\nertag}[1]{\texttt{#1}}
\newcommand{\sysname}{\textsc{StructShot}}

\aclfinalcopy 
\def\aclpaperid{2799} 

\title{Simple and Effective Few-Shot Named Entity Recognition\\ with Structured Nearest Neighbor Learning}

\author{Yi Yang \\
	ASAPP Inc.\\
	New York, NY 10007\\
	{\tt yyang@asapp.com} \\\And 
	Arzoo Katiyar \thanks{ \hspace{0.15cm}Work done at ASAPP Inc.} \\
	Pennsylvania State University \\
	University Park, PA 16802\\
	{\tt arzoo@psu.edu}
    }

\date{}

\begin{document}
\maketitle
\begin{abstract}


We present a simple few-shot named entity recognition (NER) system 
based on nearest neighbor learning and structured inference. 
Our system uses a supervised NER model trained on the source domain, as a feature extractor.
Across several test domains, we show that a nearest neighbor classifier in this feature-space is far more effective than the standard meta-learning approaches. 
We further propose a cheap but effective method to capture the label dependencies between entity tags without expensive CRF training. 
We show that our method of combining structured decoding with nearest neighbor learning achieves state-of-the-art performance on standard few-shot NER evaluation tasks, improving F1 scores by $6\%$ to $16\%$ absolute points over prior meta-learning based systems.

\end{abstract}

\section{Introduction}
\label{sec:intro}

Named entity recognition (NER) aims at identifying and categorizing spans of text into a closed set of classes, such as people, organizations, and locations. 
As a core language understanding task, NER is widely adopted in several domains, such as news~\cite{tjong2003introduction}, medical~\cite{stubbs2015annotating}, and social media~\cite{derczynski2017results}. 
However, one of the primary challenges in adapting NER to new domains is the mismatch between the different domain-specific entity types.
For example, only two out of the twenty-three entity types annotated in the I2B2 2014~\cite{stubbs2015annotating} data can be found in the OntoNotes 5~\cite{weischedel2013ontonotes} annotations. 
Unfortunately, obtaining NER annotations for a novel domain can be quite expensive, often requiring domain knowledge.

Few-shot classification~\cite{matching-net,Bao2020Few-shot} models aim at recognizing new classes based on only few labeled examples (support set) from each class.  
In the context of NER, these few-shot classification methods can enable rapid building of NER systems for a new domain by labeling only a few examples per entity class.
Several previous studies \citep{few-shot-NER,label-dependency} propose using prototypical networks~\cite{proto-net}, a popular few-shot classification algorithm,
to address the few-shot NER problem. However, these approaches 
only achieve $10 \sim 30\%$ F1 scores on average, when transferring knowledge between different NER datasets with one or five shot examples, warranting more effective methods for the problem.

\begin{figure}[t]
\centering
\includegraphics[scale=.31]{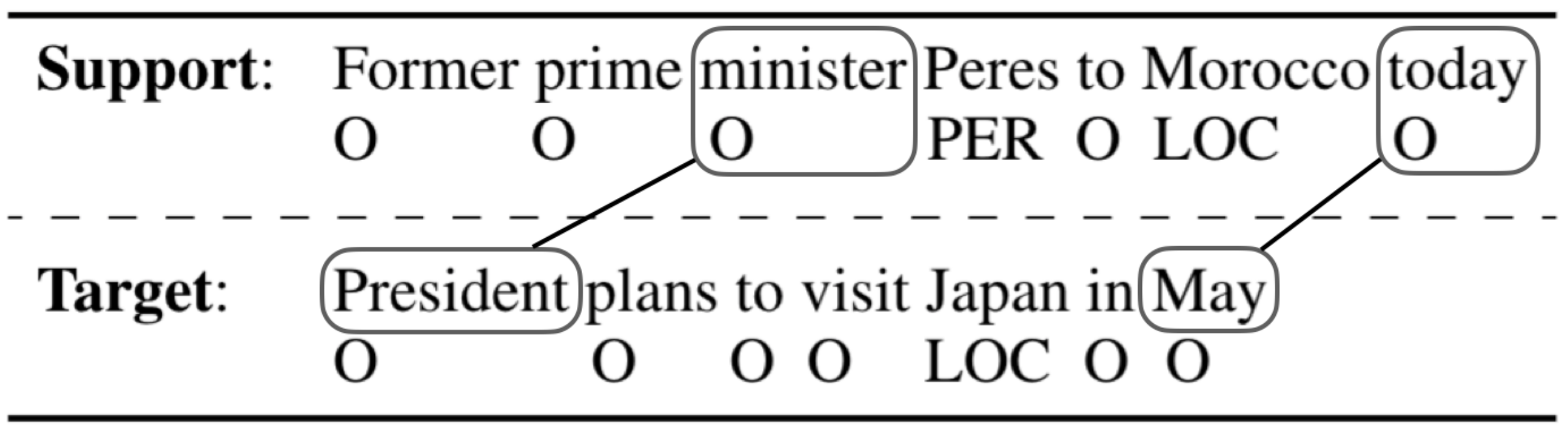}
\caption{An few-shot NER example. Professions (e.g., `minister' and `president') and dates (e.g., `today' and `tomorrow') are part of the \nertag{O} class. Nearest neighbor classifier is better at predicting the \nertag{O} class using an instance-based metric compared to methods using class-based metric.}
\label{fig:example}
\end{figure}

The direct adaption of existing few-shot \emph{classification} methods to few-shot NER is challenging for two reasons. First, NER is essentially a structured learning problem. It is crucial to model label dependencies as shown in \citet{lample2016neural} instead of directly classifying each token independently using the existing few-shot classification approaches.
Second, 
few-shot classification models \cite{proto-net} typically learn to represent each semantic class by a prototype based on the labeled examples in its support set.
However, for NER, unlike the entity classes, the \nertag{Outside} (\nertag{O}) class does not represent any unified semantic meaning. In fact, tokens labeled with \nertag{O} in a dataset actually correspond to different semantic spaces that should be separately represented in a metric-based learning framework. Consider, for example, in~\autoref{fig:example}, semantic classes such as professions (e.g., `minister') and dates (e.g., `today') may also belong to the \nertag{O} class for some NER datasets. Thus, previous approaches end up learning a noisy prototype for representing \nertag{O} class in this low-resource setting.

In this paper, we propose a simple, yet effective method~\sysname~for few-shot NER. Instead of learning a prototype for each entity class, we represent each token in the labeled examples of the support set by its contextual representation in the sentence. We learn these contextual representations from training a standard supervised NER model \cite{lample2016neural, bert}, on the source domain.
Whereas meta-learning approaches \cite{proto-net, matching-net} simulate few-shot evaluation setup during training, our approach does not need to do so. This makes it possible to deploy a unified NER system supporting both classical and emerging types of entities,
without the overhead of maintaining a separate few-shot system.
During evaluation, \sysname~ uses a nearest neighbor (NN) classifier and a Viterbi decoder for prediction. 
As shown in ~\autoref{fig:example}, for each token (``president'') in the target example, the NN classifier finds its nearest token (``minister'') from the support examples, instead of relying on an erroneous class-level (\nertag{Outside}) prototypical representation. We also improve our nearest neighbor predictions by using Viterbi decoder~\cite{forney1973viterbi} to capture label dependencies. 


We perform extensive in-domain and out-of-domain experiments for this problem. We test our systems on both identifying new types of entities in the source domain as well as identifying new types of entities in various target domains in one-shot and five-shot settings. In addition to the previous evaluation setup followed by \newcite{label-dependency}, we propose a more standard and reproducible evaluation setup for few-shot NER by using standard test sets and development sets from benchmark datasets of several domains. In particular, we sample support sets from the standard development set and evaluate our models on the standard test set.
For all our experiments, we find that our proposed systems outperform previous meta-learning systems by $6\%$ to $16\%$ absolute F1 score.


\section{Problem Statement and Setup}
\label{sec:problem}

In this section, we formalize the task of few-shot NER and propose a standard evaluation setup to facilitate meaningful comparison of results for future research.

\subsection{Few-shot NER}

NER is a sequence labeling task, where each token in a sentence is either labeled as part of an entity class (e.g., \nertag{Person}, \nertag{Location}, and \nertag{Organization}) or \nertag{O} class if it does not belong to an entity. In practice, tagging schemes such as BIO or IO are adopted to represent if a token is at the beginning (\nertag{B-X}) or inside (\nertag{I-X}) of an entity \nertag{X}. Few-shot NER focuses on a specific NER setting where a system is trained on annotations of one or more source domains $\{\mathcal{D}_\mathcal{S}^{(i)}\}$ and then tested on one or more target domains $\{\mathcal{D}_\mathcal{T}^{(i)}\}$ by only providing a few labeled examples per entity class. It is a challenging problem since the target tag set $\mathcal{C}_\mathcal{T}^{(i)}$ can be different from any source tag set $\mathcal{C}_\mathcal{S}^{(j)}$. To this end, few-shot NER systems need to learn to generalize to unseen entity classes using only a few labeled examples. 

Formally, the task of $K$-shot NER is defined as follows: given an input sentence $\mathbf{x} = \{x_t\}_{t=1}^T$ and a $K$-shot support set for the target tag set $\mathcal{C}_\mathcal{T}$, find the best tag sequence $\mathbf{y} = \{y_t\}_1^T$ for $\mathbf{x}$. The $K$-shot support set contains $K$ entity examples (not tokens) for each entity class given by $\mathcal{C}_\mathcal{T}$.

\subsection{A standard evaluation setup}

Prior work \cite{few-shot-NER, label-dependency} on few-shot NER followed few-shot classification literature and adopted the episode evaluation methodology. Specifically, a NER system is evaluated with respect to multiple evaluation episodes. An episode includes a sampled $K$-shot support set of labeled examples and a few sampled $K$-shot test sets. 
In addition to these prior practices, we propose a more realistic evaluation setting by sampling only the support sets and testing the model on the standard test sets from NER benchmarks.


\paragraph{Test set construction}
In the episode evaluation setting, test sets are sampled such that the different entity classs are equally distributed. This evaluation setup clearly does not account for the entity distributions in the real data.\footnote{In the I2B2 test data, more frequent \nertag{DATE} entity occurs 4,983 times, whereas less frequent \nertag{EMAIL} entity occurs only once.} As a result, the reported performance scores do not reflect the effectiveness of these models when adapting to a new domain. We propose to use the original test sets of the standard NER datasets to evaluate the performance of our models. Our evaluation setup does not need to randomly sample test sets, thus, improving its reproducibility for future research.


\paragraph{Support set construction}
In order to test our models in the few-shot setting, we sample support sets from the standard development set of the benchmark dataset. We account for the variance of our model performance by sampling multiple support sets and reporting the average performance on the test set for these sampled support sets. We plan to release the different support sets used for evaluation in our experiments for reproducibility.

Unlike classification tasks, a sentence in NER may contain multiple entity classes. Thus, simply sampling $K$ sentences for each entity class will result in many more entities of frequent classes than those of less frequent classes, as sampling entities of infrequent classes is more likely to also bring in entities of frequent classes than the other way around.
Because of this, we utilize a greedy sampling strategy to build support sets as shown in~\autoref{algo:support}. In particular, we sample sentences for entity classes in an increasing order with respect to their frequencies.



\begin{algorithm}[t]
\caption{Greedy sampling}
\small
\label{algo:support}
\begin{algorithmic}[1]
\REQUIRE \# of shot $K$, labeled set $\mathbf{X}$ with tag set $\mathcal{C}$
\STATE Sort classes in $\mathcal{C}$ based on their freq. in $\mathbf{X}$
\STATE $S \gets \phi $ //Initialize the support set
\STATE $\{ \text{Count}_i \gets 0 \}$ //Initialize counts of entity classes in $\mathcal{S}$
\WHILE {$i < |\mathcal{C}|$}
\WHILE{$\text{Count}_i < K$}
\STATE Sample $(\mathbf{x},\mathbf{y}) \in \mathbf{X}$ s.t. $\mathcal{C}_i \in \mathbf{y}$,  w/o replacement
\STATE $S \gets S \cup \{ (\mathbf{x},\mathbf{y}) \}$
\STATE Update $\{ \text{Count}_j \}$ $\forall \mathcal{C}_j \in \mathbf{y}$
\ENDWHILE
\ENDWHILE
\RETURN $\mathcal{S}$
\end{algorithmic}
\end{algorithm}

\section{Model}
\label{sec:model}

In this section, we present our 
few-shot NER algorithm based on structured nearest neighbor learning (\sysname). Our method uses a NER model \cite{lample2016neural,bert} trained on the source domain, as a token embedder to generate contextual representations for all tokens.
At inference, these \emph{static} representations are simply used for nearest neighbor token classification. We also use a Viterbi decoder to capture label dependencies by leveraging tag transitions estimated from the source domain.

\subsection{Nearest neighbor classification for few-shot NER}

The backbone of \sysname~is a simple token-level nearest neighbor classification system (NNShot). At inference, given a test example $\mathbf{x} = \{ x_t \}_1^T$ and a $K$-shot entity support set $\mathcal{S} = \{ (\mathbf{x}_n^{(sup)}, \mathbf{y}_n^{(sup)} \}_{n=1}^N$ comprising of $N$ sentences, NNShot employs a token embedder $f_\theta (x) = \hat{x}$ to 
obtain contextual representations for all tokens in their respective sentences. NNShot simply computes a similarity score between a token $x$ in the test example and all tokens $\{ x' \}$ in the support set. It assigns the token $x$ a tag $c$ corresponding to the most similar token in the support set:
\begin{align}
y^* &= \arg \min_{c \in \{1, \cdots, C\}} d_c(\hat{x}) \\
d_c(\hat{x}) &= \min_{x' \in \mathcal{S}_c} d(\hat{x}, \hat{x'}), \notag 
\end{align} 
where $\mathcal{S}_c$ is the set of support tokens whose tags are $c$. In this work, we use the squared Euclidean distance, $d(\hat{x}, \hat{x'}) = ||\hat{x} - \hat{x'}||_2^2$ for computing similarities between tokens in the nearest neighbor classification. We also perform L2-normalization on the features before computing these distances.


\paragraph{Pre-trained NER models as token embedders}

Most meta-learning approaches \cite{proto-net,label-dependency} simulate the test time setup during training. Hence, these approaches sample multiple support sets and test sets from the training data and learn representations to minimize their corresponding few-shot loss on the source domain. In this paper, we instead use a NER model trained on the source domain to learn token-level representations that minimizes the supervised cross-entropy loss.
Supervised NER models typically consist of a token embedder $f_\theta (\cdot)$ followed by a linear classifier $\mathbf{W} \in \mathbb{R}^{D \times L}$ where $D$ is the token embedding size and $L$ represents the number of tags. 

We consider two popular neural architectures for our supervised NER model: a BiLSTM NER model~\cite{lample2016neural} and a BERT-based NER model~\cite{bert}.\footnote{We fine-tune the cased BERT-base model.} For training these models on the source domain, we follow the setting from their original papers. These models are trained to minimize the cross-entropy loss $\ell (\mathbf{W} f_\theta(x), y)$ on the training data in the source domain. \footnote{If training data from more source domains is available, a similar multitask loss can be adopted.} At inference time, NNShot uses the BiLSTM and Transformer encoders just before the final linear classification layers as token embedders.

\subsection{Structured nearest neighbor learning}

Conditional random field (CRF)~\cite{lafferty2001conditional} is the de facto method to model label dependencies for NER. \newcite{lample2016neural} use BiLSTM embedder followed by a classification layer to represent token-tag emission scores and learn tag-tag transition scores by joint training a CRF layer.
Adopting a similar method is challenging in the context of few-shot learning. The mismatch between the tags in the source domain and the target domain does not allow learning tag-tag transition scores of the target domain by only training on the source domain. 


\begin{figure}[t]
\centering
\includegraphics[scale=.235]{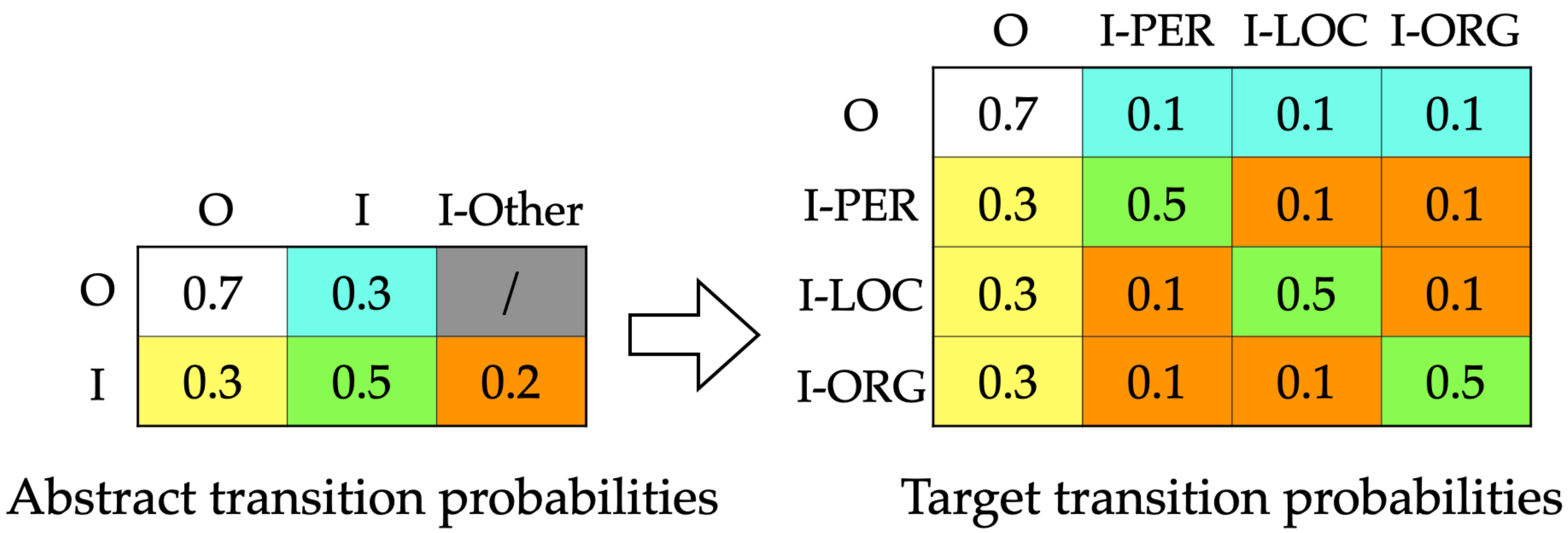}
\caption{A depiction of the extension of an abstract transition matrix. An abstract transition probability is evenly split into related target transitions, which is illustrated using the cells of the same color in their corresponding rows of the two matrices.}
\label{fig:trans}
\end{figure}

\sysname~addresses this challenge by using an abstract tag transition distribution estimated on the source domain data. Additionally, \sysname~discards training phase in CRF and only makes use of its Viterbi decoder during inference. In particular, similar to~\newcite{label-dependency}, we utilize a transition matrix that captures transition probabilities between three abstract NER tags: \nertag{O}, \nertag{I}, \nertag{I-Other}\footnote{We demonstrate the transitions with the IO tagging scheme and ignore \nertag{START} and \nertag{END} tags for simplicity.}. For instance, $p(\nertag{O} | \nertag{I})$ and $p(\nertag{I} | \nertag{O})$ correspond to the transition probabilities between an entity tag and \nertag{O}, whereas  $p(\nertag{I} | \nertag{I})$ and $p(\nertag{I-Other} | \nertag{I})$ correspond to the probabilities of transitioning from an entity tag to itself and to a different entity tag respectively. As depicted in~\autoref{fig:trans}, we can extend these abstract transition probabilities to an arbitrary target domain tag set by evenly distributing the abstract transition probabilities into corresponding target transitions. Our simple extension method guarantees that the resulting target transition probabilities still lead to a valid distribution. \newcite{label-dependency} copy these abstract transition scores to multiple specific transitions such that the resulting target transition probabilities no longer correspond to a distribution. 

The key idea in \sysname~is that it estimates the abstract transition probabilities by counting the number of times a particular transition was observed in the training data. 
The transition probability from \nertag{X} to \nertag{Y} is
\begin{equation}
    p(\nertag{Y} | \nertag{X}) = \frac{N (\nertag{X} \rightarrow \nertag{Y})}{N (\cdot \rightarrow \nertag{Y})},
\end{equation}
where $N (\nertag{X} \rightarrow \nertag{Y})$ and $N (\cdot \rightarrow \nertag{Y})$ are the frequencies of the transition from $\nertag{X}$ to $\nertag{Y}$ and the transition from any tag to $\nertag{Y}$ respectively. In practice, these abstract transitions can also be drawn from a prior distribution given domain knowledge.

For Viterbi inference, we obtain the emission probabilities  $p(y = c | x)$  for each token in the test example from NNShot. 
\begin{equation}
    p(y = c | x) = \frac{e^{- d_c (\hat{x})}}{\sum_{c'} e^{- d_{c'} (\hat{x})}}.
\end{equation}
Given this abstract transition distribution $p(y' | y)$ and the emission distribution $p(y | x)$, we use Viterbi decoder to solve the following the structured inference problem:
\begin{equation}
    \mathbf{y}^* = \arg\max_{\mathbf{y}} \prod_{t=1}^T p(y_t | x) \times p (y_t | y_{t-1}).
\end{equation}
As the emission and transition probabilities are estimated independently, we introduce a temperature hyper-parameter $\tau$ that re-normalizes the transition probabilities to align the emission and transition scores to a similar scale.


\section{Experiments}
\label{sec:exp}

In this section, we compare \sysname~against existing methods on two few-shot NER scenarios: tag set extension and domain transfer. We adopt several benchmark NER corpora in different domains for the few-shot experiments.\footnote{When ready, the code will be published at \url{https://github.com/asappresearch/structshot}.}

\subsection{Data}
\begin{table} [ht!]
\centering
\small
\begin{tabular}{lrrrr}
    \toprule
     Dataset &  Domain & \# Class & \# Sent & \# Entity \\ \midrule
     OntoNotes & General & 18 & 76,714 & 104,151  \\
     CoNLL'03 & News & 4 & 20,744 & 35,089  \\
     I2B2'14 & Medical & 23 & 140,817 & 29,233 \\
     WNUT'17 & Social & 6 & 5,690 & 3,890 \\
    \bottomrule
\end{tabular}
\caption{Data statistics. \# Class corresponds to the number of entity classes labeled in a dataset.}
\label{tab:data}
\end{table}

We experiment with standard NER datasets in four important domains: OntoNotes 5.0~\cite{weischedel2013ontonotes} (General), CoNLL 2003~\cite{tjong2003introduction} (News), I2B2 2014~\cite{stubbs2015annotating} (Medical), and WNUT 2017~\cite{derczynski2017results} (Social). To the best of our knowledge, these are the largest annotated NER corpora in their respective domains. These datasets are labeled with diverse and representative named entity types.
\autoref{tab:data} presents detailed statistics of these datasets. We use the OntoNotes train/development/test splits released for the CoNLL 2012 shared task.\footnote{Available at: \url{http://conll.cemantix.org/2012/data.html}} Standard train/development/test splits also come with other dataset distributions. 

\subsection{Evaluation tasks}
We evaluate few-shot NER systems on two real world scenarios. For both scenarios, we experiment with both one-shot and five-shot settings.

\paragraph{Tag set extension}
Our first set of experiments are motivated by the fact that new types of entities often emerge in some domains such as medical and social media. Thus, we evaluate the performance of our systems on recognizing new entity types as they emerge in the source domain. We mimic this scenario by splitting the entity classes of a dataset into a source set and a target set. Specifically, we randomly split the eighteen entity classes of the OntoNotes dataset into three target entity class sets:
\begin{itemize}
\small
    \item Group A: \{\nertag{ORG}, \nertag{NORP}, \nertag{ORDINAL}, \nertag{WORK\_OF\_ART}, \nertag{QUANTITY}, \nertag{LAW}\}
    \item Group B: \{\nertag{GPE}, \nertag{CARDINAL}, \nertag{PERCENT}, \nertag{TIME}, \nertag{EVENT}, \nertag{LANGUAGE}\}
    \item Group C: \{\nertag{PERSON}, \nertag{DATE}, \nertag{MONEY}, \nertag{LOC}, \nertag{FAC}, \nertag{PRODUCT}\}
\end{itemize}
We evaluate our systems on each target entity set. For each experiment, we modify the training set by replacing all the entity tags corresponding to the target test group with the \nertag{O} tag. Hence, these target tags are no longer observed during training.
Similarly, we modify the test set to only include annotations corresponding to the target test group such that we only evaluate our models based on the unseen tags during training. As discussed in~\autoref{sec:problem}, we sample multiple support sets from the development set to simulate the few-shot setting at the test time.

\paragraph{Domain transfer}
The second set of experiments address a common scenario of adapting a NER system to a novel domain. For our experiments, we use General (OntoNotes) as the source domain and test our models on News (CoNLL), Medical (I2B2) and Social (WNUT) domains. We train our supervised NER models on the standard OntoNotes training set, whereas we evaluate the few-shot systems on standard test sets of CoNLL, I2B2, and WNUT. The support sets are sampled from the corresponding development sets of the three corpora.

\subsection{Experimental settings}

We have provided details of our proposed evaluation setup for few-shot NER in ~\autoref{sec:problem}. 
We report the standard evaluation metrics for NER: micro averaged F1 score. For each experiment, we sample five support sets and report the mean and standard deviation of the corresponding F1 scores. In order to establish a comprehensive comparison with prior work, we also report episode evaluation results in the Appendix. 

\paragraph{Competitive systems}
We consider five competitive approaches in our experiments. We build BERT-based systems for all the methods and BiLSTM-based systems for three of them.
\textit{Prototypical Network}~\cite{proto-net} is a popular few-shot classification algorithm that has been adopted in most state-of-the-art (SOTA) few-shot NER systems~\cite{few-shot-NER}. \textit{PrototypicalNet+P\&D}~\cite{label-dependency} improves upon Prototypical Network by using the pair-wise embedding and dependency transfer mechanism.\footnote{\newcite{label-dependency} show that Matching Network~\cite{matching-net} preforms worse than Prototypical Network on their evaluation for few-shot NER.} \textit{SimBERT} is a nearest neighbor classifier based on the pre-trained BERT encoder without fine-tuning on any NER data. Finally, we include our proposed \textit{NNShot} and \textit{\sysname} described in~\autoref{sec:model}. We use the IO tagging scheme for all of the experiments, as we find that it performs much better than BIO scheme for all the considered methods.

\begin{table*} [ht!]
\centering
\small
\begin{tabular}{lllllllll}
    \toprule
    \multirow{2}{*}{System} & \multicolumn{4}{c}{Tag Set Extension} & \multicolumn{4}{c}{Domain Transfer} \\
    \cmidrule(l){2-5} \cmidrule(l){6-9}
          & Group A & Group B & Group C & Ave. & CoNLL & I2B2 & WNUT & Ave. \\ \midrule
   \multicolumn{9}{l}{\it BiLSTM-based systems} \\
     Prototypical Network & \: $4.0{\pm}1.6$  & \: $5.4{\pm}1.9$ & \: $5.2{\pm}1.5$ & \: $4.9$ & $18.7{\pm}9.2$  & \: $2.2{\pm}1.0$ & \: $5.5{\pm}2.7$ & \: $8.8$ \\
     NNShot (ours) & $15.7{\pm}7.1$  & $25.1{\pm}7.1$ & $22.7{\pm}7.1$  & $21.2$ & $46.4{\pm}11.7$ & \: $7.5{\pm}2.9$ & \: $6.9{\pm}3.2$ & $20.3$ \\
     \sysname~(ours) & $18.9{\pm}9.4$ & $31.9{\pm}5.1$ & $22.0{\pm}3.4$ & $24.3$ & $53.1{\pm}9.9$ & $10.5{\pm}2.6$ & $10.4{\pm}4.4$ & $24.7$ \\[6pt]
   \multicolumn{9}{l}{\it BERT-based systems}\\
     SimBERT  & \: $8.3{\pm}1.4$ & \: $9.0{\pm}3.8$ & \: $8.4{\pm}1.8$ & \: $8.6$ & $15.7{\pm}3.7$ & \: $7.7{\pm}0.8$ & \: $4.9{\pm}1.2$ & \: $9.4$ \\
     Prototypical Network  & $18.7{\pm}4.7$ & $24.4{\pm}8.9$ & $18.3{\pm}6.9$ & $20.5$ & $53.0{\pm}7.2$ & \: $7.6{\pm}3.5$ & $14.8{\pm}4.9$ & $25.1$ \\
     PrototypicalNet+P\&D & $18.5{\pm}4.4$ & $24.8{\pm}9.3$ & $20.7{\pm}8.4$ & $21.3$ & $56.0{\pm}7.3$ & \: $7.9{\pm}3.2$ & $18.8{\pm}5.3$ & $27.6$ \\
     NNShot (ours) & $27.2{\pm}3.5$ & $\textbf{32.5}{\pm}14.4$ & $\textbf{23.8}{\pm}10.2$ & $27.8$ & $61.3{\pm}11.5$ & $16.6{\pm}2.1$ & $21.7{\pm}6.3$ & $33.2$ \\
     \sysname~(ours) & $\textbf{27.5}{\pm}4.1$ & $32.4{\pm}14.7$ & $\textbf{23.8}{\pm}10.2$ & $\textbf{27.9}$ & $\textbf{62.3}{\pm}11.4$ & $\textbf{22.1}{\pm}3.0$ & $\textbf{25.3}{\pm}5.3$ & $\textbf{36.6}$  \\
    \bottomrule
\end{tabular}
\caption{F1 score results on one-shot NER for both tag set extension and domain transfer tasks. We report standard deviations from runs with five different support sets sampled from the validation sets. The best results are in {\bf bold}.}
\label{tab:1shot-results}
\end{table*}

\begin{table*} [ht!]
\centering
\small
\begin{tabular}{llllllllll}
    \toprule
    \multirow{2}{*}{System} & \multicolumn{4}{c}{Tag Set Extension} & \multicolumn{4}{c}{Domain Transfer} \\
    \cmidrule(l){2-5} \cmidrule(l){6-9}
          & Group A & Group B & Group C & Ave. & CoNLL & I2B2 & WNUT & Ave. \\ \midrule
   \multicolumn{9}{l}{\it BiLSTM-based systems} \\
     Prototypical Network & \: $7.4{\pm}2.7$  & $21.8{\pm}7.6$ & $18.2{\pm}5.6$ & $15.8$ & $47.6{\pm}9.0$ & \: $5.9{\pm}1.1$ & \: $8.8{\pm}3.3$ & $20.8$ \\
     NNShot (ours) &  $24.5{\pm}5.4$ & $35.2{\pm}7.4$ & $33.8{\pm}6.3$ & $31.2$  & $62.0{\pm}6.1$ & \: $8.4{\pm}2.7$ & $12.4{\pm}4.2$ & $27.6$ \\
     \sysname~(ours) & $26.1{\pm}6.0$ & $46.1{\pm}6.5$ & $38.0{\pm}1.8$ & $36.7$ & $63.8{\pm}6.9$ &  $13.7{\pm}0.8$ & $15.1{\pm}4.9$ & $30.9$ \\[6pt]
   \multicolumn{9}{l}{\it BERT-based systems}\\
     SimBERT  & $10.1{\pm}0.8$ & $23.0{\pm}6.7$ & $18.0{\pm}3.5$ & $17.0$ & $28.6{\pm}2.5$ & \: $9.1{\pm}0.7$ & \: $7.7{\pm}2.2$ & $15.1$ \\
     Prototypical Network  & $27.1{\pm}2.4$ & $38.0{\pm}5.9$ & $38.4{\pm}3.3$ & $34.5$ & $65.9{\pm}1.6$ & $10.3{\pm}0.4$ & $19.8{\pm}5.0$ & $32.0$ \\
     PrototypicalNet+P\&D & $29.8{\pm}2.8$ & $41.0{\pm}6.5$ & $38.5{\pm}3.3$ & $36.4$ & $67.1{\pm}1.6$ & $10.1{\pm}0.9$ & $23.8{\pm}3.9$ & $33.6$  \\
     NNShot (ours) & $44.7{\pm}2.3$ & $53.9{\pm}7.8$ & $53.0{\pm}2.3$ & $50.5$ & $74.3{\pm}2.4$ & $23.7{\pm}1.3$ & $23.9{\pm}5.0$ & $40.7$ \\
     \sysname~(ours) & $\textbf{47.4}{\pm}3.2$ & $\textbf{57.1}{\pm}8.6$& $\textbf{54.2}{\pm}2.5$ & $\textbf{52.9}$ & $\textbf{75.2}{\pm}2.3$ & $\textbf{31.8}{\pm}1.8$ & $\textbf{27.2}{\pm}6.7$ & $\textbf{44.7}$  \\
    \bottomrule
\end{tabular}
\caption{F1 score results on five-shot NER for both tag set extension and domain transfer tasks. We report standard deviations from runs with five different support sets sampled from the validation sets. The best results are in {\bf bold}.}
\label{tab:5shot-results}
\end{table*}

\paragraph{Parameter tuning}
We adopt the best hyperparameter values reported by~\cite{yang2018design} for the BiLSTM-NER models and use the default BERT hyper-parameter values provided by Hugging Face\footnote{\url{https://huggingface.co/}}. Specifically, our BiLSTM-NER models adopt one-layer word-level BiLSTM model and one-layer character-level uni-directional LSTM model. LSTM hidden sizes are $50$ and $200$ and input embedding sizes are $30$ and $100$ for the character-level and word-level models respectively. We use the pre-trained $100$-dimensional GloVe vectors~\cite{pennington2014glove} to initialize the word embeddings for all BiLSTM-NER models. SGD and Adam~\cite{adam} are utilized to optimize the BiLSTM-based and BERT-based models with learning rates $0.015$ and $5 \times 10^{-5}$ respectively. We tune other parameters required by different few-shot learning methods on the source domain development sets. The transition re-normalizing temperature $\tau$ is chosen from $\{ 0.01, 0.005, 0.001\}$.

\subsection{Results}


The results for one-shot NER and five-shot NER are summarized in~\autoref{tab:1shot-results} and~\autoref{tab:5shot-results} respectively. As shown, our NNShot and \sysname~perform significantly better than all previous methods across all evaluation settings. 
By modeling label dependencies with a simple Viterbi decoder, \sysname~boosts the performance of NNShot by $2.4\%$ and $4\%$ F1 scores on five-shot tag set extension and domain transfer tasks on average respectively. These performance gains are greater than the ones obtained by joint CRF training with the prototypical network (PrototypicalNet+P\&D), suggesting that independently modeling transition and emission scores is a cheap but effective way to capture label dependencies. \sysname~achieves new SOTA results on the two few-shot NER tasks, outperforming the previous SOTA system (PrototypicalNet+P\&D) by $6\%$ to $9\%$ F1 score on one-shot setting and $11\%$ to $16\%$ F1 score on five-shot setting.

\paragraph{BiLSTM vs. BERT as token embedder} The BERT-based systems considerably outperform BiLSTM-based systems on few-shot NER. Language model pre-training is critical for low-resource natural language processing including few-show transfer learning~\cite{colin2019proceedings}. However, task-specific knowledge is usually more important than the general information learned via unsupervised training. For example, the top-performing BiLSTM-based systems can beat SimBERT by up to $15\%$ F1 score on some few-shot NER settings. With fine-tuning on the OntoNotes data, NNShot outperforms SimBERT by $20\%$ to $35\%$ F1 scores across different settings, demonstrating the effectiveness of injecting task-specific information into pre-trained language models. 

\paragraph{Tag set extension vs. Domain transfer} The one-shot NER systems generally perform better on domain transfer than on tag set extension, while the five-shot systems work better on the tag set extension task. On the domain transfer task, the source entity classes overlap with some entity classes in the target domain, which benefits NER systems built under the extremely low-resource condition. However, in general, domain transfer is more challenging than tag set extension due to language variation across different domains. Not surprisingly, our five-shot NER systems are not only more accurate but also more robust than the one-shot systems. The standard deviations reported with multiple five-shot support sets are much lower than those obtained with one-shot support sets. This indicates that we can build more reliable few-shot NER systems given more few-shot examples in the support sets.

\paragraph{Episode evaluation} Finally, as shown in the Appendix, the results obtained on episode evaluation are generally better than the ones reported with our proposed evaluation setup. However, the performance trend is the same, i.e.,  \sysname~significantly outperforms all competitors. It implies that previous studies \cite{few-shot-NER,label-dependency} overestimate the performance of their few-shot NER systems. 

\paragraph{Few-shot NER in practice} Although the average F1 scores of the few-shot NER systems are relatively low, we believe that few-shot NER systems are still very useful in practice. First, the few-shot NER results are reasonably good if the source and target domains are close to each other. For example, the five-shot NER system trained on the OntoNotes training set can achieve 75\% F1 score on the CoNLL test set. Second, given the few-shot NER system, we are able to provide immediate support to emerging entity types without retraining and redeploying the NER model. At the same time, a more accurate NER model can be trained in parallel after collecting sufficient annotations for the new types. 

\subsection{Analysis}

We perform analysis to investigate the impact of various tagging schemes and BERT fine-tuning objectives on few-shot NER.

\paragraph{Tagging scheme}
When only a few entities are available in the support sets, the conventional BIO tagging scheme can harm the performance of few-shot NER systems, as it further reduces the number of labeled instances per tag class. We experiment with both BIO and IO tagging schemes for all the few-shot NER models. The systems equipped with IO tagging scheme always outperform those with BIO scheme. In particular, \sysname~and NNShot benefit from switching from BIO scheme to IO scheme by an average of $3.2\%$ and $3.8\%$ F1 scores on the five-shot tag set extension and domain transfer tasks respectively.

\paragraph{Fine-tuning objective}
\sysname~exploits the standard cross-entropy loss for NER used in the original paper~\cite{bert} to fine-tune BERT on OntoNotes data. 
We also experiment with fine-tuning BERT using the prototypical network objective, and then utilize the encoder in \sysname. The results show that BERT fine-tuned with the standard NER loss performs much better than the one fine-tuned with the prototypical network loss by $12\%$ and $9\%$ on the five-shot tag set extension and domain transfer tasks respectively. 
This suggests that the popular meta-learning methods fall short in capturing effective representations for few-shot NER task. 



\section{Discussion}
\label{sec:discuss}

In this section, we investigate two questions: 1) why~\sysname~is so effective? and 2) why few-shot NER is so difficult?

\begin{figure}[t]
\centering
\includegraphics[scale=.24]{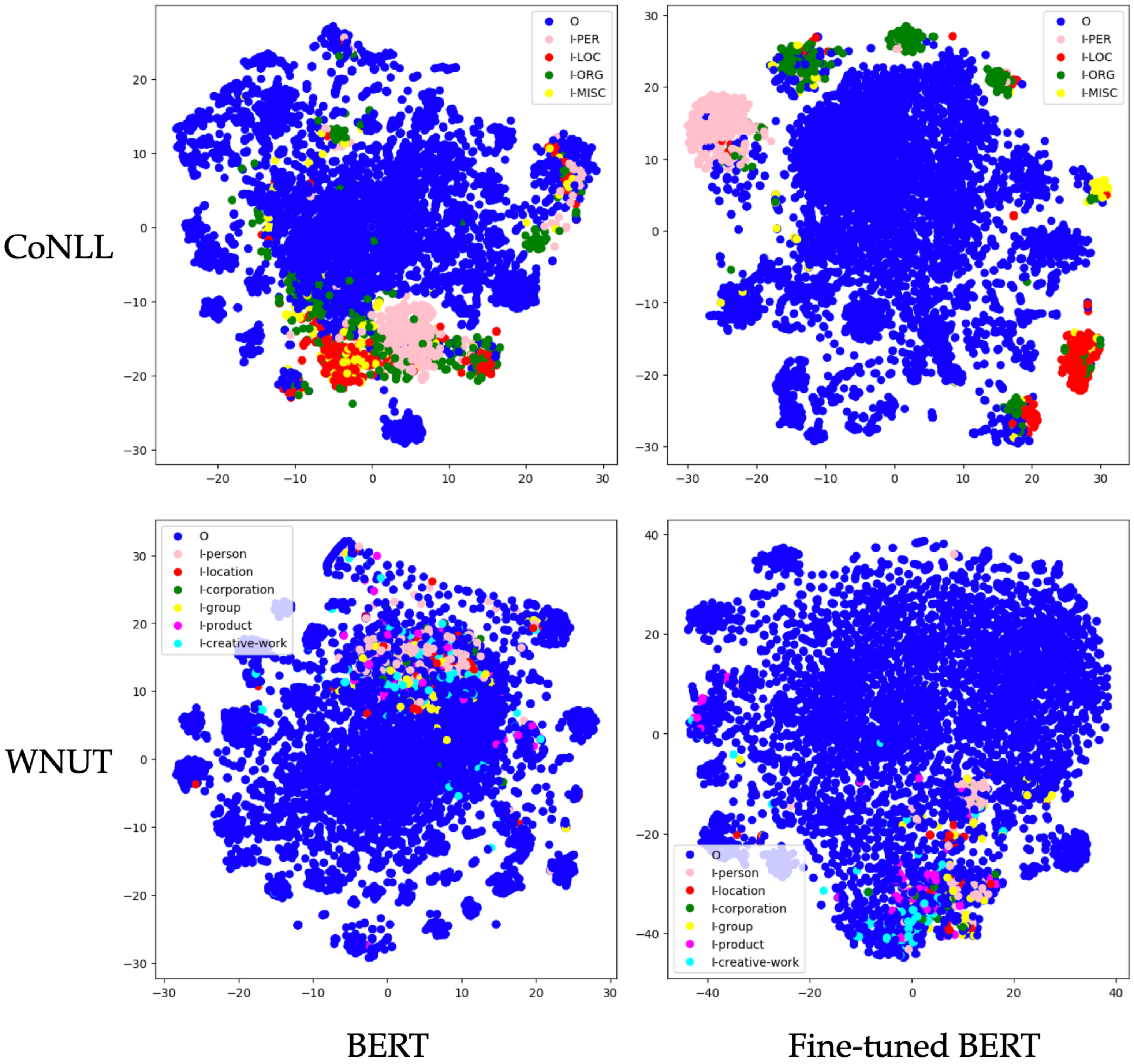}
\caption{t-SNE visualizations of the CoNLL and WNUT test sets. The representations are obtained from the pre-trained BERT-base model and the BERT-base model fine-tuned on the OntoNotes training data.}
\label{fig:tsne}
\end{figure}

\paragraph{t-SNE visualization} We project token-level representations obtained from the BERT embedders onto a 2-dimensional space using t-SNE~\cite{maaten2008visualizing}. \autoref{fig:tsne} presents the visualization results on the CoNLL and WNUT test sets (we exclude I2B2 as it includes too many classes for visualization).
 Fine-tuning BERT on OntoNotes clearly improves the task-awareness with respect to both CoNLL and WNUT datasets, as instances of the same class are much closer compared to those obtained from the non-fine-tuned BERT model. The separation of different entity classes is more evident on CoNLL due to the greater tag set overlap with OntoNotes. 
 Instances labeled with \nertag{O} are spread across the space, regardless of fine-tuning. 
 This explains the effectiveness of \sysname. First, fine-tuning BERT in a conventional NER setting is able to learn a good entity specific metric space. Second, the nearest neighbor classifier that emphasizes more on local distance is more appropriate for assigning \nertag{O} to an instance. 

\begin{table} [t!]
\centering
\small
\addtolength{\tabcolsep}{-2pt}
\begin{tabular}{lrrlrr}
    \toprule
     \multicolumn{3}{c}{I2B2} & \multicolumn{3}{c}{WNUT}\\
    \cmidrule(l){1-3} \cmidrule(l){4-6}
      Class  & F1 & \# Entity & Class & F1 & \# Entity \\ 
    \cmidrule(l){1-3} \cmidrule(l){4-6}
    \nertag{DATE} & 66.1 & 4,983 & \nertag{person} & 57.8 & 429 \\
    \nertag{CITY} & 61.1 & 260 & \nertag{loc.} & 49.6 & 150 \\
    \nertag{DOCTOR} & 51.1 & 1,945 & \nertag{c-work} & 30.1 & 142 \\
    \nertag{AGE} & 29.3 & 768 & \nertag{corp.} & 16.7 & 66 \\
    \nertag{MED-REC} & 14.4 & 428 & \nertag{product} & 10.9 & 127 \\
    \nertag{PATIENT} & 14.0 & 920 & \nertag{group} & 5.0 & 165 \\
    \nertag{HOSPITAL} & 10.1 & 874 & - & - & - \\
    \nertag{PHONE} & 7.9 & 224 & - & - & - \\
    \nertag{IDNUM} & 7.0 & 201 & - & - & - \\
    \bottomrule
\end{tabular}
\caption{Best per-class five-shot domain transfer results obtained from \sysname~on the I2B2 and WNUT test sets, in which \nertag{MED-REC}, \nertag{loc.}, \nertag{c-work}, and \nertag{corp.} correspond \nertag{MEDICAL-RECORD}, \nertag{location}, \nertag{creative-work}, and \nertag{corporation} respectively.}
\label{tab:error-analysis}
\end{table}

\paragraph{Per-class performance analysis} We attempt to shed some light on the second question by analyzing outputs from the best five-shot \sysname~systems on the domain transfer task. The per-class F1 scores are shown in~\autoref{tab:error-analysis}, where we exclude I2B2 classes with less than 200 instances in the test set. \sysname~achieves reasonable performance on less ambiguous entity classes such as \nertag{DATE}, \nertag{CITE}, \nertag{person}, and \nertag{location}. However, it struggles to distinguish between highly ambiguous classes. For example, \nertag{AGE}, \nertag{MEDICAL-RECORD}, \nertag{PHONE}, and \nertag{IDNUM} are all numbers. It is still challenging for our system to differentiate different numerical types without any domain specific knowledge. Similarly, \sysname~often predicts a \nertag{PATIENT} entity as \nertag{DOCTOR} and it nearly always assigns the \nertag{corporation} label to entities of \nertag{group}. We believe that domain specific cues like `Dr.' and `MD.' can be useful in resolving these ambiguities and enable few-shot NER systems to generalize better.


\section{Related Work}
\label{sec:related}

\paragraph{Meta learning} Meta learning is widely studied in the computer vision community, as the low-level features in images are transferable across classes that enables learning from only a few examples from the unseen class. The existing approaches \cite{proto-net, matching-net} typically focus on metric learning.
\newcite{proto-net} learn a prototype representation for each class and classifies test points based on the nearest prototypes. \newcite{matching-net} compute support set aware similarities between a test point and the target classes. These methods have been adapted with some success to NLP tasks including text classification \citep{yu-etal-2018-diverse, geng-etal-2019-induction, Bao2020Few-shot}, machine translation~\cite{gu-etal-2018-meta}, and relation classification~\cite{han-etal-2018-fewrel}. Recently, \citet{simple-shot} show that simple feature transformations followed by nearest neighbor search can perform competitively with the state-of-the-art meta-learning methods on standard computer vision classification datasets. Inspired by this approach, we evaluate the performance of nearest neighbor based classification against meta-learning methods.

\paragraph{Few-shot NER} A few approaches have been proposed for few-shot NER. \newcite{hofer2018few} explore different pre-training and fine-tuning strategies to recognize entities in medical text with a few examples. \newcite{few-shot-NER} and~\newcite{label-dependency} exploit popular few-shot classification methods such as prototypical networks and matching network, where~\newcite{label-dependency} also jointly learn transition scores that improve performance. These approaches require complex episode training and only achieve unsatisfactory results. \sysname~ does not require meta-training. With a simple nearest neighbor classifier and a structured decoder, it is much more accurate than other existing meta-learning based systems.
\section{Conclusion}
\label{sec:con}
We introduce \sysname, a simple few-shot NER system that achieves SOTA performance without any few-shot specific training. We identify two weaknesses of previous systems related to their handling of \nertag{O} class and modeling label dependencies. Our systems overcomes these challenges with nearest neighbor learning and structured decoding. We further propose a standard evaluation setup for few-shot NER and show that \sysname~significantly outperforms prior SOTA systems on popular benchmarks across multiple domains. In the future, we want to extend our system to other few-shot sequence tagging problems such as part-of-speech tagging and slot filling.

\section*{Acknowledgments}
We thank the EMNLP reviewers for their
helpful feedback. We also thank the ASAPP NLP  
team for their support throughout the project.

\bibliographystyle{acl_natbib}
\bibliography{cite-strings,cites,cite-definitions}

\newpage
\begin{appendices}
\section{Appendix: Episode Evaluation Results}
\label{app:mappings}

The main episode evaluation results for one-shot NER and five-shot NER are summarized in~\autoref{tab:1shot-results-epi} and~\autoref{tab:5shot-results-epi} respectively. We sample $100$ evaluation episodes for each experiment. The performance trend is the same as our main results, in which \sysname~significantly outperforms all competitors. 

\vspace{.2cm}

\begin{table*} [ht!]
\centering
\small
\begin{tabular}{lllllllll}
    \toprule
    \multirow{2}{*}{System} & \multicolumn{4}{c}{Tag Set Extension} & \multicolumn{4}{c}{Domain Transfer} \\
    \cmidrule(l){2-5} \cmidrule(l){6-9}
          & Group A & Group B & Group C & Ave. & CoNLL & I2B2 & WNUT & Ave. \\ \midrule
   \multicolumn{9}{l}{\it BiLSTM-based systems} \\
     Prototypical Network & \: $4.0{\pm}1.6$ & \: $5.4{\pm}1.9$ & \: $5.1{\pm}1.5$ & \: $4.8$ & $15.3{\pm}7.5$ & \: $3.6{\pm}1.3$ & \: $2.6{\pm}1.4$ & \: $7.2$ \\
     NNShot (ours) & $15.7{\pm}7.0$ & $27.5{\pm}7.1$ & $20.1{\pm}6.0$ & $21.1$ & $49.6{\pm}12.6$ & \: $9.5{\pm}4.1$ & \: $8.9{\pm}1.6$ & $22.7$ \\
     \sysname~(ours) & $18.9{\pm}9.7$ & $32.0{\pm}5.1$ & $22.0{\pm}3.3$ & $24.3$ & $50.0{\pm}9.2$ & $11.0{\pm}2.0$ & \: $9.9{\pm}4.2$ & $23.6$ \\[6pt]
   \multicolumn{9}{l}{\it BERT-based systems}\\
     SimBERT  & $13.0{\pm}1.8$ & $14.3{\pm}3.9$ & \: $9.5{\pm}1.1$ & $12.3$ & $19.3{\pm}4.3$ & $16.3{\pm}2.1$ & \: $5.3{\pm}0.9$ & $13.6$ \\
     Prototypical Network  & $25.5{\pm}3.7$ & $30.5{\pm}6.8$ & $21.2{\pm}5.8$ & $25.7$ & $59.3{\pm}6.3$ & $19.9{\pm}2.7$ & $15.8{\pm}4.1$ & $31.6$ \\
     PrototypicalNet+P\&D & $27.2{\pm}1.1$ & $31.4{\pm}6.9$ & $23.0{\pm}5.1$ & $27.2$ & $61.7{\pm}6.8$ & $21.3{\pm}4.8$ & $17.5{\pm}2.9$ & $33.5$ \\
     NNShot (ours) & $\textbf{31.3}{\pm}4.5$ & $32.8{\pm}7.4$ & $ 27.3{\pm}7.8$ & $30.5$ & $67.6{\pm}10.8$ & $30.1{\pm}2.5$ & $20.2{\pm}6.0$ & $39.3$ \\
     \sysname~(ours) & $30.8{\pm}5.0$ & $\textbf{33.5}{\pm}7.7$ & $\textbf{28.0}{\pm}7.9$ & $\textbf{30.8}$ & $\textbf{68.7}{\pm}10.5$ & $\textbf{32.1}{\pm}1.7$ & $\textbf{20.5}{\pm}5.2$ & $\textbf{40.4}$  \\
    \bottomrule
\end{tabular}
\caption{F1 score results of episode evaluation on one-shot NER for both tag set extension and domain transfer tasks. We report standard deviations from runs with five different support sets sampled from the validation sets. The best results are in {\bf bold}.}
\label{tab:1shot-results-epi}
\end{table*}

\begin{table*} [ht!]
\centering
\small
\begin{tabular}{llllllllll}
    \toprule
    \multirow{2}{*}{System} & \multicolumn{4}{c}{Tag Set Extension} & \multicolumn{4}{c}{Domain Transfer} \\
    \cmidrule(l){2-5} \cmidrule(l){6-9}
          & Group A & Group B & Group C & Ave. & CoNLL & I2B2 & WNUT & Ave. \\ \midrule
   \multicolumn{9}{l}{\it BiLSTM-based systems} \\
     Prototypical Network  & \: $7.4{\pm}2.7$ & $23.9{\pm}6.2$ & $18.2{\pm}5.6$ &  $16.5$ &  $49.2{\pm}5.8$ & \: $8.5{\pm}4.6$  & \: $5.2{\pm}1.8$  &  $21.0$ \\
     NNShot (ours)  & $24.5{\pm}5.8$ & $42.3{\pm}12.9$ & $33.8{\pm}6.3$ &  $33.5$ & $62.1{\pm}6.8$ & $12.4{\pm}4.2$  & \: $9.0{\pm}2.6$  &  $27.8$ \\
     \sysname~(ours) & $26.1{\pm}6.0$ & $47.0{\pm}7.7$ & $38.0{\pm}1.8$ & $37.1$ & $63.8{\pm}6.9$ & $13.7{\pm}0.8$  & $15.1{\pm}4.9$ & $30.9$ \\[6pt]
   \multicolumn{9}{l}{\it BERT-based systems}\\
     SimBERT  & $18.8{\pm}1.4$ & $27.0{\pm}2.4$ & $21.4{\pm}3.1$ & $22.4$ & $31.7{\pm}1.3$ & $23.6{\pm}1.7$ & \: $9.3{\pm}2.2$ & $21.6$ \\
     Prototypical Network  & $36.4{\pm}0.9$ & $46.3{\pm}2.0$ & $41.6{\pm}1.4$ & $41.4$ & $69.2{\pm}2.0$ & $27.6{\pm}2.8$ & $22.1{\pm}3.1$ & $39.6$ \\
     PrototypicalNet+P\&D & $38.5{\pm}4.1$ & $49.5{\pm}2.3$ & $44.3{\pm}1.2$ & $44.1$ & $69.6{\pm}2.3$ & $32.2{\pm}2.1$ & $26.0{\pm}2.1$ & $42.6$  \\
     NNShot (ours) & $45.3{\pm}1.5$ & $53.4{\pm}2.9$ & $49.9{\pm}1.3$ & $49.5$ & $77.2{\pm}1.8$ & $45.4{\pm}2.1$ & $26.7{\pm}4.0$ & $49.8$ \\
     \sysname~(ours) & $\textbf{47.2}{\pm}0.9$ & $\textbf{54.9}{\pm}2.9$& $\textbf{51.2}{\pm}1.4$ & $\textbf{51.1}$ & $\textbf{77.9}{\pm}1.8$ & $\textbf{46.1}{\pm}3.2$ & $\textbf{27.9}{\pm}3.2$ & $\textbf{50.6}$  \\
    \bottomrule
\end{tabular}
\caption{F1 score results of episode evaluation on five-shot NER for both tag set extension and domain transfer tasks. We report standard deviations from runs with five different support sets sampled from the validation sets. The best results are in {\bf bold}.}
\label{tab:5shot-results-epi}
\end{table*}

\end{appendices}

\end{document}